# Stable and Faithful Explanations for Knowledge Tracing

Praveena Padi [1], Arun Morampudi [2], Ujval Sai Gopal Irrinki [3], and Pradeep Kumar Dolabehera Kakitapelli [4]

[1] Georgia Institute of Technology, Atlanta, Georgia, USA

[2] Independent Researcher, Atlanta, Georgia, USA

[3] Amazon Web Services, Austin, Texas, USA

[4] University of the Cumberlands, Kentucky, USA

Corresponding author: Pradeep Kumar Dolabehera Kakitapelli (pkakitapelli58721@ucumberlands.edu)

## Abstract

Knowledge tracing (KT) models predict student performance opaquely, limiting pedagogical action. This study contributes a validation protocol testing predictive competitiveness (RQ1), explanation stability (RQ2) and retraining-based faithfulness (RQ3) together. Thirteen behavioral features across five pedagogical themes were engineered from ASSISTments 2009 and 2012, with history features computed from temporally preceding interactions and current response latency retained only for retrospective analysis. ASSISTments 2009 was rebuilt: the uncorrected skill-builder release duplicates each multi-skill interaction across one row per skill, and because those rows share one correctness label, they leak it into preceding-interaction features. Rebuilding lowered model AUC and reordered the explanation results. An Extreme Gradient Boosting (XGBoost) model explained with Tree SHapley Additive exPlanations (TreeSHAP) was compared against four deep baselines (DKT, SAKT, AKT and SimpleKT) under an information-matched protocol giving the deep models the same behavioral signals and restricting XGBoost to what is derivable from the identifier-and-correctness stream they consume. XGBoost reached an area under the curve (AUC) of 0.777 on 2012 and 0.786 on rebuilt 2009, with prediction-time AUCs of 0.771 and 0.775, respectively, after excluding current response latency; restricted to the baselines' information it performed as they did (0.697 against 0.700, and 0.717 against 0.720), locating the difference in information supplied, not model family. Rankings were consistent across folds, seeds and conditioning schemes (Spearman ρ = 0.989–1.000), and removing top-ranked TreeSHAP features harmed AUC more than random removal, though split-gain and permutation rankings performed comparably. Student-level examples are illustrative interpretations, not validated recommendations.

## 1. Introduction

Knowledge tracing (KT), the task of modeling a student's evolving knowledge state to predict performance on future learning items, has become a central component of intelligent tutoring systems [1]. Deep learning approaches, from Deep Knowledge Tracing (DKT) [2] to attention-based architectures such as Self-Attentive Knowledge Tracing (SAKT) [3], Attentive Knowledge Tracing (AKT) [4], and SimpleKT [5], achieve strong predictive accuracy but function as black boxes. Without understanding why a student is predicted to fail, instructors cannot translate the model outputs into targeted interventions.

This gap is not merely practical in nature. Learning science frameworks, particularly self-regulated learning theory [6] and the knowledge-learning-instruction framework [7], emphasize that effective instructional support requires the identification of the specific processes (e.g., spacing, mastery accumulation, response latency) that drive student outcomes. An opaque probability score does not provide such process-level insights.

Recent studies have applied post-hoc explainability in educational contexts. SHapley Additive exPlanations (SHAP) have been used to identify predictors of course outcomes [8]. However, systematic validation of explanation stability across folds, seeds, and TreeSHAP conditioning schemes, and faithfulness testing through ablation, remain limited in KT literature. This study supplies validation, and states explicitly what each check does and does not license. Because the intended readership includes learning scientists and practitioners who do not work in explainable artificial intelligence (XAI), Section 3.2 introduces the attribution and evaluation machinery used here from first principles before it is applied.

### Research Questions

RQ1 (Accuracy): Under an information-matched protocol, does a feature-based model that admits post-hoc explanation match the predictive accuracy of deep KT architectures on standard benchmark datasets?

RQ2 (Stability): Is the global feature-importance ranking consistent across cross-validation folds, random seeds, and the two structurally different conditioning schemes available for TreeSHAP?

RQ3 (Faithfulness): Do the features that TreeSHAP ranks highest measurably influence model predictions, in the sense that removing them and retraining degrades held-out performance more than removing an equal number of features chosen at random, and how does the TreeSHAP ranking compare with established alternative importance measures under the same test?

### Related Work

Knowledge-tracing models. KT has evolved from Bayesian Knowledge Tracing (BKT) [9] to deep architectures. DKT [2] applied long short-term memory networks (LSTMs) to sequential interactions; SAKT [3] introduced self-attention, allowing flexible weighting of past interactions; AKT [4] incorporated Rasch-inspired embeddings with monotonic attention; and SimpleKT [5] showed that a simplified attention mechanism achieves competitive performance. These four were selected to span the three architectural families that dominate the deep-KT literature rather than to represent the state of the art: a recurrent family (DKT), a plain self-attention family (SAKT), and a family that augments attention with explicit item-difficulty parameters (AKT and SimpleKT, which differ in whether monotonic attention and a context-aware distance decay are used). Section 3.1 states the shared configuration and the per-model input schema. Deep architectures that consume behavioral and temporal signals directly, including DKT-Forget [48], LPKT [49], DIMKT [50] and LBKT [51], form a distinct family and are discussed in Section 3.1 as information-matched comparators. Feature-based approaches have remained competitive: logistic models with well-crafted features can match deep KT on several benchmarks [10], and feature-rich models incorporating response time and practice history perform strongly in student modeling [11].

Post-hoc explainability is also important. SHAP [12] provides theoretically grounded attribution based on the Shapley values. TreeSHAP [13] enables the exact computation of tree ensembles. For neural networks, Integrated Gradients [14] provides axiomatically justified attributions. The faithfulness of attention weights as explanations remains a topic of debate [15, 16]. Beyond SHAP, LIME [17] provides model-agnostic local explanations through surrogate linear models, although the instability of its explanations across repeated sampling runs has been documented [52, 53]. A broad taxonomy distinguishes intrinsically interpretable models from post-hoc attribution techniques [19, 54]. That distinction is central here and is used strictly: a gradient-boosted ensemble of several hundred trees is not intrinsically interpretable, and this study does not claim it is. What XGBoost offers over a deep sequence model in this setting is a *tabular input space of named, semantically meaningful features* over which exact Shapley values can be computed in polynomial time [13], which makes both global attribution and feature-removal experiments tractable. Separately, Slack et al. [18] showed that post-hoc explainers including LIME and SHAP can be deliberately manipulated by an adversarial model; that finding motivates auditing explanations rather than trusting them, but it concerns adversarial construction and not the run-to-run variability studied here.

Explainable artificial intelligence (XAI) validation and faithfulness. A growing body of work addresses whether explanations genuinely reflect the behavior of models. Robustness metrics for explanation stability demonstrate that small input perturbations can yield large attribution changes in some methods [20]. RemOve and Retrain (ROAR) was proposed as a faithfulness benchmark on the grounds that feature removal without retraining confounds feature importance with distribution shift [21]. Comprehensiveness and sufficiency were formalized as complementary faithfulness metrics in the ERASER benchmark for NLP rationale extraction [22]. Extensions to multi-class settings show that faithfulness varies substantially across explanation methods and tasks [23].

XAI in education. A survey of explainable AI in education noted the need for validation beyond single-model and single-dataset demonstrations [24]. Feature-based approaches have also been explored in student modeling [25]. Foundational guidelines for educational data mining (EDM) methodology emphasize the importance of interpretable models for practitioner adoption [26]. A survey of two decades of EDM research identified explanation quality as a persistent gap between research and classroom deployment [27]. The present study extends this body of work by providing multi-dataset and multi-method stability and faithfulness validation of global feature rankings for knowledge tracing.

Explainability specific to knowledge tracing. Work explaining KT models specifically, as opposed to student-success prediction generally, falls into three groups. The first makes the model itself transparent: Interpretable Knowledge Tracing [55] replaces the latent state with named learner variables, and BKT [9] is interpretable by construction at the cost of per-skill independence. The second applies post-hoc attribution to a deep KT model, using layer-wise relevance propagation on DKT [56] or reading attention weights, an approach whose status as explanation is contested [15, 16]. The third, which includes this study, uses a feature-based model with pedagogically named inputs and attributes predictions to them [10, 11, 57]. Prior work in the third group generally reports an importance ranking and stops there; this study treats the ranking as an empirical claim, measuring it for stability across folds, seeds and conditioning schemes, and for faithfulness against non-random competing rankings under retraining. Swamy et al. [58] found that explanation methods disagree and that instructors do not uniformly trust them, a further reason to characterize stability before proposing classroom use.

Foundations of learning science. The features in this study were grounded in established learning science constructs. Self-regulated learning (SRL) theory [6, 28] emphasizes that learners' metacognitive monitoring, reflected in response timing and practice patterns, drives performance outcomes. A review of SRL assessment instruments noted that behavioral traces (timestamps and accuracy sequences) serve as unobtrusive SRL indicators [29]. The spacing effect, which states that distributed practice improves retention over massed practice, is a well-established finding in learning science [30, 31], providing a theoretical justification for the Recency/Spacing features in this study. Item difficulty calibration draws on item response theory (IRT) [32], where the Rasch model formalizes difficulty as a latent parameter estimated from the response patterns.

## 2. Data and Features

### 2.1 Datasets

Dataset statistics are summarized in Table 1.

Table 1. Dataset statistics

| Dataset | Interactions | Users | Skills | Mean Seq. Len | Median Seq. Len | Correctness Rate |
|---|---|---|---|---|---|---|
| ASSISTments 2012 | 2,568,649 | 18,022 | 265 | 142.5 | 79.0 | 0.697 |
| ASSISTments 2009 | 241,212 | 2,062 | 107 | 117.0 | 56.5 | 0.659 |

ASSISTments 2012 serves as the primary dataset with real wall-clock timestamps [33, 34]. ASSISTments 2009 provides a contrasting dataset that differs in scale, skill granularity, and temporal resolution [35]. Both were cleaned to a canonical schema (`user_id`, `skill_id`, `problem_id`, correct ∈ {0,1}, timestamp, `response_time_ms`), sorted by (`user_id`, timestamp), with users having fewer than 20 interactions excluded. The exact preprocessing sequence for each dataset is given in Section 2.2, and the corresponding code path is named there so that every step can be in the released repository.

ASSISTments 2009 lacks wall-clock timestamps; therefore, time-gap features were computed as positional differences. The rankings are reported with and without these features for transparency.

Both datasets originated from the same platform. While they differ substantially in scale (six times more interactions in 2012), temporal resolution, student population, and skill granularity (265 vs. 107 skills), they do not constitute evidence of generalization to other platforms (e.g., EdNet, Junyi Academy). The second dataset tests whether the findings are replicated under different data characteristics within the same platform.

### 2.2 Preprocessing Protocol

Preprocessing runs to completion before any model is fitted and is independent of model training and inference. It is implemented in `src/xaikt/data.py` and is invoked by `scripts/01_prepare_data.py`; the cleaned corpora are cached and reused unchanged by every downstream stage, so no partition of the data can influence how another partition is cleaned.

ASSISTments 2012. The source file is *2012-2013-data-with-predictions-4-final.csv*, obtained from the Figshare record cited in the Data Availability statement. Six columns are read (`user_id`, `skill_id`, `problem_id`, *correct*, `start_time`, `ms_first_response`). Rows with a missing value in any of these are dropped; rows whose *correct* value is not in {0, 1} are dropped; `start_time` is parsed to Unix seconds; identifiers are remapped to contiguous zero-based integers; rows are sorted by (`user_id`, *timestamp*, original file order); and users with fewer than 20 remaining interactions are excluded. This release records one row per interaction. Auditing the cleaned corpus for repeated observations found 303 rows sharing a (`user_id`, *timestamp*) pair, 0.01% of the corpus, and no rows at all sharing a (`user_id`, *timestamp*, `problem_id`) triple, so no deduplication step is required.

ASSISTments 2009. This study uses *skill_builder_data.csv* from the uncorrected 2009–2010 skill-builder release, in which an interaction tagged with more than one skill is written as one row per skill, those rows sharing a single `order_id`, a single `problem_id` and a single *correct* label. Xiong et al. [59] documented the artifact and showed that removing the duplicates moves reported DKT performance from approximately 0.86 to the 0.73–0.75 range; the pyKT benchmark [60] treats deduplication as required. Because the features here summarise the interactions preceding the current one, and duplicate rows sort adjacently, a duplicate copy places the interaction's own label inside its own history window. Of the 459,208 source rows carrying a skill tag, only 283,105 are distinct interactions; the remaining 176,103 are repeated copies. Before deduplication, `rolling_acc_k` alone separated the correct from the incorrect responses of the duplicate copies at an AUC of 0.965, against 0.674 on first copies, the signature of a feature reading its own label.

The corpus was rebuilt as follows. Rows with a missing `skill_id` are dropped. Scaffolding rows are removed by retaining only rows with *original* = 1, so each row is a main problem rather than a tutoring sub-question. Interactions flagged under multiple skills are reduced to one row, keyed on `order_id`, retaining the tagged skill with the highest corpus frequency and breaking ties by the student's most recently practiced skill, the one-skill-per-row convention used by the competition split and by pyKT [60]. Rows whose *correct* value is not in {0, 1} are dropped, `order_id` is retained as the ordering key, identifiers are remapped to contiguous zero-based integers, rows are sorted by (`user_id`, `order_id`), and users with fewer than 20 remaining interactions are excluded. The scaffolding filter removes 26,047 rows and deduplication a further 173,762, leaving 259,399 distinct interactions before the minimum-length filter. The rebuilt corpus contains 241,212 interactions from 2,062 students over 107 skills, against 443,338 interactions from 2,324 students over 123 skills previously; Table 1 gives the full statistics. All ASSISTments 2009 results in this manuscript were recomputed on it.

No feature encodes the current response, and three checks are reported. First, by construction: each feature is emitted by a single forward pass over a student's ordered interactions before the current row's label enters the running state (`src/xaikt/features.py`), so no state visible at row t has observed the label at row t. The two difficulty features are the exception and are handled in Section 2.4. Second, empirically: the univariate AUC of each feature against the current label is computed as a screen, since a feature encoding the current response would separate the classes far better than any summary of prior behavior. On the rebuilt corpus the largest value is 0.697 (`prior_acc_skill`), followed by `rolling_acc_k` at 0.677 and by `problem_difficulty` and `prior_acc_overall` at 0.674, with every remaining feature at 0.620 or below. This screen is what exposed the defect described above, where the value on duplicate copies was 0.965. Third, the faithfulness experiment acts as a detector in its own right; Section 5 describes an earlier target-encoding defect caught that way.

### 2.3 Behavioral Features

Thirteen features were engineered across five themes, listed in Table 2. History features are computed from interactions that strictly precede the interaction being predicted for that student; current response latency and difficulty encoding are treated separately in Sections 2.5 and 2.4. This is a temporal constraint on feature construction, not a causal claim: it rules out one specific form of information leakage, and it does not establish that any feature exerts a causal influence on learning. The manuscript uses the phrase *temporally preceding* for this property throughout.

From interaction sequences to tabular instances. Both model families predict whether the current response is correct. The tabular model receives one row per cleaned interaction, with history features summarizing that student's earlier interactions; the deep baselines consume the sequence directly (Section 3.1). No windowing, subsampling or sequence truncation is applied to the tabular data. The accumulated-accuracy features (prior_acc_skill, prior_acc_overall and rolling_acc_k) are strictly past-only: they are computed within each student in order_id sequence on 2009 and timestamp order on 2012, before the current label enters the running state.

Table 2. Behavioral features and thematic groups

| Theme | Features | Description |
|---|---|---|
| Latency | response_time_log, rolling_mean_response_time | Current and recent response speed. response_time_log is excluded from the prediction-time input set (Section 2.5). |
| Prior Mastery | prior_acc_skill, prior_attempts_skill, prior_acc_overall, rolling_acc_k | Accumulated correctness, practice volume, and accuracy over the ten most recent interactions (k = 10). rolling_acc_k was reassigned from Recency/Spacing in this revision (Section 5). |
| Recency/ Spacing | gap_since_last, gap_since_skill | Elapsed time since the previous interaction and since the previous interaction on the same skill. Positional proxies on ASSISTments 2009. |
| Difficulty | skill_difficulty, problem_difficulty | Fold-internal target encoding of item and skill easiness, computed from training rows only (Section 2.4). |
| Engagement/ Position | seq_position, attempts_in_session, session_index | Position in the learning sequence and session structure. |

Features that require at least one earlier interaction are undefined at the point where that history does not yet exist (rolling_acc_k at a student's first interaction, prior_acc_skill at the first encounter with a skill), and are emitted as NaN rather than imputed with a value that would imply a history the student does not have. XGBoost handles NaN natively via its learned default-direction splits [36], and Logistic Regression receives median-imputed values.

Difficulty features use K-fold internal target encoding ($K$ = 5) to prevent the data leakage. A naive leave-one-out (LOO) correction was initially attempted but introduced a worse leak: the LOO value becomes a deterministic two-valued function of each row's label, which tree models memorize perfectly (test AUC dropped from ~0.77 to ~0.63; skill_difficulty inflated to 36% of total SHAP importance). K-fold encoding addresses both self-inclusion and deterministic function leakage. This bug was detected by the faithfulness check (Section 4.3), which failed before the fix and passed afterward, providing a concrete demonstration of faithfulness testing as a pipeline validation.

Several features correlate substantially with one another, as behavioral traces drawn from a single interaction stream inevitably do. Of the seventy-eight feature pairs, ten reach a Spearman correlation of 0.40 or above in absolute value on ASSISTments 2012 and nine do so on the rebuilt ASSISTments 2009 corpus, with a similar structure on both: position with session index (ρ = 0.92 and 0.65), the two elapsed-gap features with each other (0.77 and 0.84), the accumulated-accuracy features with rolling accuracy (0.64 and 0.70, 0.57 and 0.65) and with each other (0.51 and 0.58), the two latency features (0.47 on both), and the two difficulty features (0.45 on both). The complete matrix for both corpora is produced by the released code. This correlation structure is a property of the domain and cannot be removed; Section 3.3 explains what it does and does not imply for the attribution results, and why reporting two conditioning schemes is a diagnostic rather than a remedy.

### 2.4 Difficulty Encoding, Unseen Identifiers, and Session Boundaries

Difficulty encoding summarizes how often a skill or problem was answered correctly, without letting a row's own label determine its input. Both features use only the outer fold's training rows (`src/xaikt/features.py`, called per fold from `scripts/02_train_models.py`); validation and test labels never enter the aggregates. Within training, rows are assigned to five internal groups using fixed seed 42. Each row receives the mean correctness for its `skill_id` or `problem_id` from the other four groups. Validation and test rows receive means from the full training partition. No additive smoothing or minimum-count threshold is used. An identifier absent from the source rows receives `NaN`, which XGBoost handles through its learned default split direction [36]. The five-group count and seed 42 remain fixed across experiment seeds.

The outer training, validation and test partitions are student-disjoint. As a sensitivity check, we refitted all fifteen runs per dataset with the five internal encoding groups assigned by student (fixed seed 42), retaining the outer splits and model configuration. The aggregate path-dependent ranking was unchanged on both datasets (ρ = 1.000 versus row-wise encoding); `problem_difficulty` remained first. Mean cross-fold ρ was 0.993 and 1.000, respectively. The explanation analyses retain the original row-wise encoding results.

Session boundaries are defined by an inactivity threshold on the ordering key. For ASSISTments 2012, where wall-clock timestamps are available, a gap of more than 1,800 seconds (30 minutes) from the student's previous interaction opens a new session. For ASSISTments 2009, where only an ordering key is available, a gap of more than 20 positions serves the same role. The positional variant is a weaker proxy and is treated as such: it is a stand-in for elapsed time and is not interpreted as an interval.

### 2.5 The Prediction-Time Input Set

The prediction-time input set. `response_time_log` is the logarithm of the current interaction's first-response latency, which is not observable until the student has already responded and so is unavailable when a deployed tutoring system needs the prediction. Its use is defensible only under a retrospective framing, where the question is which behavioral signals carry information about correctness rather than what a system could act on in advance. To keep the two apart, the prediction-time input set is defined as the twelve features fully determined before the current response is submitted, and predictive performance is reported for both sets in Section 4.1; the main explanation analyses use the full retrospective set, with the twelve-feature sensitivity analysis reported in Section 4.2. Removing the feature costs little: AUC falls from 0.777 [0.775, 0.779] to 0.771 [0.769, 0.773] on ASSISTments 2012, and from 0.786 [0.779, 0.792] to 0.775 [0.769, 0.782] on ASSISTments 2009. The reduction is real on ASSISTments 2012, where the intervals do not overlap, and small enough on both that the prediction-time model has higher AUC than the deep baselines under the shared, untuned configurations evaluated here; this does not establish superiority over tuned deep KT models. `rolling_mean_response_time`, which averages latencies over earlier interactions only, remains in the prediction-time set.

## 3. Models and Methods

Fig. 1 summarizes the end-to-end study design, from the construction of temporally preceding features and predictive modeling to explanation generation and validation.

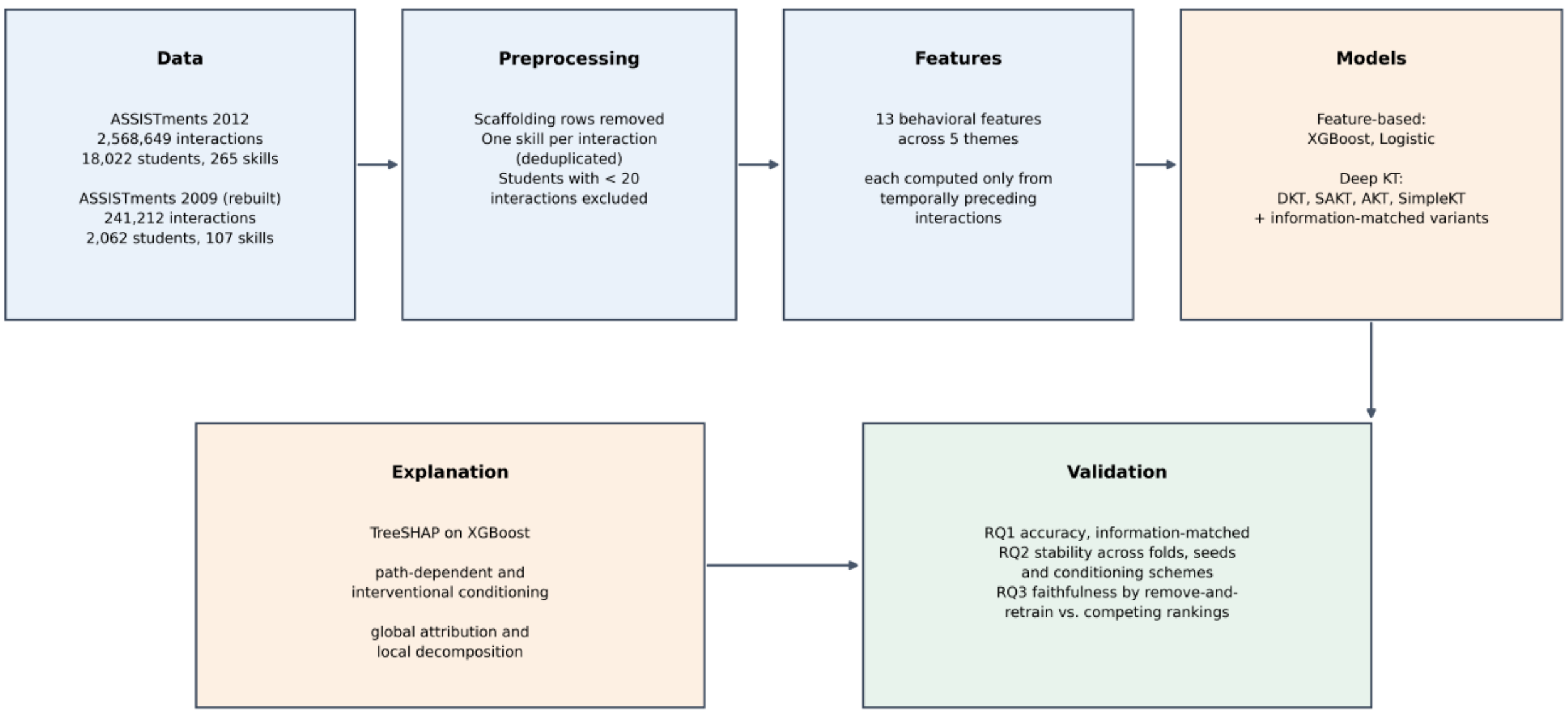

**Fig. 1** Explainable knowledge tracing pipeline

### 3.1 Models

Feature-based models. XGBoost [36] and Logistic Regression [37] are fitted on the tabular instances of Section 2.3. XGBoost uses 300 boosting rounds, maximum depth 6, learning rate 0.1, subsample and column-sample fractions of 1.0, minimum child weight 1, no L1 penalty and the default L2 penalty of 1.0, the binary logistic objective and the *hist* tree method; early stopping monitors validation AUC with patience 20 rounds, and the best round is used for test scoring. Logistic Regression uses L-BFGS with at most 1,000 iterations and median imputation for undefined values, which it cannot represent natively. Both are called *feature-based* rather than *interpretable*; Section 5 explains the distinction.

Deep KT baselines. Four architectures spanning three families were trained: DKT, a single-layer LSTM over skill-and-correctness embeddings; SAKT, multi-head self-attention over the same stream; and AKT and SimpleKT, which add question embeddings and a Rasch-style scalar difficulty per question. All four were implemented as standalone PyTorch modules rather than taken from pyKT [60], because the design required access to intermediate activations and attention tensors; Section 6 notes the reproducibility cost. Shared configuration: embedding dimension 64, hidden dimension 128, 4 attention heads, maximum sequence length 200, batch size 384, Adam at learning rate 0.001 with default β, binary cross-entropy over masked positions, at most 100 epochs, and early stopping on validation AUC with patience 10. Sequences longer than 200 interactions are split into non-overlapping windows, the first position of each excluded from the loss for want of preceding context. Training used one NVIDIA GeForce RTX 3080 with CUDA 12.1.

Input schema of the deep baselines. Tables 3 and 4 state the complete input schema of every model. The schemas are not equivalent: DKT and SAKT consume a zero-based `skill_id` embedding and a binary correctness embedding only, while AKT and SimpleKT additionally consume a zero-based `problem_id` embedding and a learned scalar difficulty per problem, so their Rasch components are active. None of the four receives response latency, inter-interaction gaps, session structure or sequence position, all of which XGBoost receives.

Table 3. Model inputs

| **Model** | **Family** | **Inputs at each step** | **Behavioral / temporal inputs** |
|---|---|---|---|
| DKT | Recurrent (LSTM) | `skill_id`, correct | None |
| SAKT | Self-attention | `skill_id`, correct | None |
| AKT | Attention + Rasch | `skill_id`, `problem_id`, correct | None |
| SimpleKT | Attention + Rasch | `skill_id`, `problem_id`, correct | None |
| DKT-Forget [48] | Recurrent + temporal | `skill_id`, correct, discretised repeated-gap and sequence-gap | Elapsed gaps |
| DKT + features | Recurrent + behavioral | `skill_id`, correct, 13 (or 12) behavioral features | All |
| SAKT + features | Self-attention + behavioral | `skill_id`, correct, 13 (or 12) behavioral features | All |
| XGBoost | Gradient-boosted trees | 13 behavioral features (12 in the prediction-time set) | All |

| Model | Family | Inputs at each step | Behavioral / temporal inputs |
|---|---|---|---|
| XGBoost (restricted) | Gradient-boosted trees | `prior_acc_skill`, `prior_attempts_skill`, `prior_acc_overall`, `rolling_acc_k`, `seq_position` | None |
| Logistic Regression | Linear | 13 behavioral features (12 in the prediction-time set) | All |

Table 4. Model input encoding

| Model | Encoding |
|---|---|
| DKT | Learned 64-d embedding of `skill_id`; learned 64-d embedding of binary correctness; the two concatenated to 128-d |
| SAKT | Learned 64-d skill and correctness embeddings plus a learned positional embedding; causal mask over 4 heads |
| AKT | Learned 64-d skill, question, and correctness embeddings, a learned scalar Rasch difficulty per question, and a positional embedding, projected to 64-d |
| SimpleKT | As AKT, with the simplified attention formulation of [5] |
| DKT-Forget [48] | As DKT, with learned embeddings of the discretised gap buckets concatenated |
| DKT + features | As DKT, plus standardized features projected by a learned linear map to 64-d and added to the step embedding |
| SAKT + features | As SAKT, plus the same learned feature projection |
| XGBoost | Raw feature values; undefined values routed by the learned default split direction |
| XGBoost (restricted) | As XGBoost, confined to features derivable from the (`skill_id`, correct) stream |
| Logistic Regression | Median imputation of undefined values; L-BFGS |

Information-matched comparison. We test whether an accuracy difference comes from the model family or from the information supplied to it. First, we give DKT and SAKT the continuous features available to XGBoost: training-partition statistics standardize them, a learned linear map projects them to the embedding dimension, and the result is added to each step embedding. Model width and the recurrent or attention mechanism remain unchanged. We also train DKT-Forget [48] to include an established architecture in this comparison. Second, we restrict XGBoost to features derived from the identifier-and-correctness stream used by DKT and SAKT: `prior_acc_skill`, `prior_attempts_skill`, `prior_acc_overall`, `rolling_acc_k` and `seq_position`. This excludes latency, gap, session and target-encoded difficulty features. Section 4.1 reports both directions of the comparison.

The deep baselines share one standardized configuration and were not tuned per model or per dataset. The comparison of interest is whether a feature-based model remains competitive when every model receives a conventional configuration, and tuning four architectures across two datasets, five folds and three seeds would multiply the training budget by the size of the search grid without changing what RQ2 and RQ3 measure. The consequence is that tuned deep models could perform better than reported here, which is one reason this manuscript claims only competitiveness and not superiority. The stability and faithfulness findings concern the explanation of a single model and are unaffected. No hyperparameter search was performed for any model, including XGBoost.

All attention-based models require a structural fix at position 0, where full causal masking leaves softmax without a valid key and produces `NaN` gradients that propagate to every shared parameter. Position 0 is allowed to self-attend, and its output is used for neither loss nor attribution; a finite-gradient assertion guards each training run.

### 3.2 Methodological Positioning

We ask two questions about an importance ranking: does it remain consistent across repeated runs, and does removing its leading features reduce predictive performance? The protocol combines established stability criteria [20], the comprehensiveness-style criterion in ERASER [22], and ROAR's remove-and-retrain procedure [21]. We adapt ROAR from image pixels to named educational features: rank the features, remove the highest-ranked ones, retrain, and measure the performance drop. Named features also allow removal by theme. The same procedure can evaluate any ranking over this feature set, allowing direct comparisons with other importance measures (Section 3.4).

Unlike ERASER, which operates on token-level rationales in sequential models, this setting has thirteen engineered features with known semantics, which enables theme-level ablation as a complementary test: if an attribution method correctly identifies Difficulty as the most important theme, removing every Difficulty feature should produce the largest single-theme AUC drop. LIME [17] is not used as an alternative explainer, because a local linear approximation is redundant for a tree ensemble whose exact Shapley values are computable, but the documented run-to-run variation in its explanations [52, 53] is part of what motivates the multi-seed, multi-fold stability protocol applied here to any attribution method.

### 3.3 Attribution Methods, Stated for Non-Specialists

SHAP [12] divides the difference between a model's prediction and a reference expectation among the input features. For one prediction, it assigns each feature a signed value: positive values raise the prediction relative to the reference, and negative values lower it. The values sum to the prediction's difference from the reference. They are Shapley values, originating in cooperative game theory [45], where credit for a joint outcome is allocated according to fairness axioms. Computing them generally requires an exponentially growing number of feature subsets. TreeSHAP [13] uses the structure of tree ensembles to compute exact values in polynomial time, making XGBoost practical for this attribution study.

Attributing a prediction to a subset of features requires deciding what the model should be taken to do about the features left out, and there is no single correct answer [61, 62]. TreeSHAP offers two conditioning schemes. Under path-dependent conditioning, omitted features are handled by following the tree structure and weighting each branch by the proportion of training data that flowed down it, so the attribution reflects the conditional distribution as represented inside the fitted trees. Under interventional conditioning, omitted features are replaced by values drawn from a background sample, so the attribution reflects an intervention that breaks the dependence between omitted and retained features [63]. The two answer different questions, roughly *true to the data* versus *true to the model* [63], and can distribute credit differently among correlated features. Neither assumes that features are independent; that description would be inaccurate for either scheme.

This matters because behavioral features in this setting are correlated, as Section 2.3 quantifies, and that correlation is a property of how students interact with a tutoring system rather than a defect a choice of estimator can remove. Reporting both conditioning schemes does not resolve it. What agreement between them supports is narrower, and is the only claim made here: the ordinal ranking of features is not an artifact of which scheme was chosen. Individual attribution magnitudes are not claimed to be invariant, and Section 5 gives an instance where they visibly are not.

A plausible explanation need not reflect what a model uses. We test faithfulness by removing the highest-ranked features, retraining the model, and comparing the loss of held-out performance with removal under competing rankings. This is the specific sense of faithfulness used throughout the manuscript. We retrain because simply setting features to zero at prediction time can create inputs unlike the training data; the resulting loss would mix the effect of missing information with the effect of unfamiliar inputs [64]. Retraining follows ROAR [21] and lets each model learn from the reduced feature set.

### 3.4 Explainability Methods and Faithfulness Tests

TreeSHAP is applied to the fitted XGBoost model under both conditioning schemes defined in Section 3.3. Path-dependent conditioning is exact and requires no background sample. Interventional conditioning draws a background sample of 1,000 training rows per fold. Global importance is the mean absolute SHAP value per feature over the explained rows, aggregated over the five folds and three seeds. Both schemes explain the same sample of up to 5,000 held-out rows per run.

Where rankings are computed, and where they are evaluated. Rankings that feed the ablation experiments are computed on the validation partition of each fold, which is student-disjoint from the test partition and is already the partition used for early stopping; the ablation is then evaluated on the untouched test partition, so the ranking and its evaluation are independent. Deriving the ranking from the test partition instead turns out to make no difference to the outcome: in the six saved seed-level comparator summaries (three seeds per dataset), the validation-derived and test-derived TreeSHAP rankings produce identical ablation AUC values at every reported value of *k*, so the ranking comparison in Section 4.4 reports one series per ranking method.

Integrated Gradients [14, 38] and attention-weight extraction were implemented for the deep baselines and are produced by the released pipeline. The quantitative comparison between those attributions and the TreeSHAP rankings that the submitted Methods section described was attempted and abandoned, for the reason given in Section 6: the only vocabulary common to a per-feature tabular attribution and a per-interaction sequence attribution was the five-theme mapping, and a rank correlation over five items is not informative. No cross-model agreement claim is therefore made, and none of this manuscript's conclusions rests on those attributions.

Stability. For each seed and each conditioning scheme, the global importance ranking is computed within every fold, the Spearman rank correlation is taken over all fold pairs, and the resulting coefficients are averaged. Seed-to-seed agreement compares the three rankings obtained by averaging feature importance over the five folds within each seed (three seed pairs), separately for each scheme. The stability analysis retains the original inter-method estimator: correlate the two fold-averaged rankings within each seed, then average the three correlations. We also report the correlation of the grand-mean rankings in Section 4.2. Top-k Jaccard overlap is the intersection size divided by the union size of the two feature sets; fold comparisons use 30 within-seed pairs per scheme, seed comparisons use the three fold-averaged seed pairs, and scheme comparisons use 15 matched runs. The feature-importance tables give descriptive sample SD across the fifteen runs, not a confidence interval.

Faithfulness. Three ablation experiments are run, all of which remove features from both the training and test matrices and refit XGBoost from scratch on the reduced feature space before scoring the held-out test partition.

The first removes the top *k* features of a given ranking, for $k \in \{1, 2, 3, 5, 7, 9\}$. The second removes *k* features drawn uniformly at random, repeated over ten independent draws, and reports the mean together with the dispersion across those draws rather than the mean alone. The third removes all features belonging to one theme at a time.

Competing rankings. Comparing a ranking against random removal establishes only that it beats chance, which is a weak bar. Three non-random rankings are therefore evaluated within the same ablation procedure: TreeSHAP mean absolute value, XGBoost split gain, and permutation importance measured on the validation partition. Reporting all three separates two distinct claims: that a reasonable importance ranking outperforms random selection, and that the TreeSHAP ranking specifically is the more faithful of the available rankings. Theme-level results are additionally normalized by the number of features removed, since the themes are of unequal size and an unnormalized comparison rewards the larger theme.

### 3.5 Experimental Protocol

Splitting and cross-validation. To evaluate predictions for unseen students, each run assigns 70% of students to training, 10% to validation and 20% to testing. All interactions from one student stay together. For each base seed, five runs shuffle the full student list using the base seed plus the fold index. We call these runs folds, but they are repeated random subsamples, not disjoint parts of a conventional k-fold partition. Their test sets overlap because each samples the same student population. The fifteen fold-by-seed runs therefore are not independent observations; Section 3.6 accounts for this in uncertainty reporting.

Seeds and training-run count: Three base seeds (0, 1, 2) are used. Each seed seeds the Python, NumPy and PyTorch generators and, offset by the fold index, that fold's student shuffle; the internal target-encoding split uses its own fixed constant of 42, as Section 2.4 notes. The revised study comprises 11 model variants × 3 seeds × 5 folds × 2 datasets = 330 model fits, plus the ablation refits. Each fold-and-seed combination contributes 90 XGBoost refits (4 rankings (validation TreeSHAP, test TreeSHAP, split gain and permutation) × 6 values of k, 6 × 10 random draws, 5 theme ablations and 1 unablated baseline), giving 1,350 per dataset and 2,700 in total, for 3,030 fits in the main experiments. Second-revision diagnostics add 30 grouped-encoding refits and two representative learning-curve refits. The released code reproduces this count. The seed + `fold_index` split rule reuses some outer partitions across base seeds; seed stability therefore measures the stated repeated-split protocol, not independent replicate datasets.

Metrics: The primary metric is the area under the receiver operating characteristic curve (AUC). Because the local examples in Section 4.5 rest on absolute predicted probabilities rather than on rankings alone, accuracy, root mean squared error, area under the precision–recall curve and expected calibration error are reported for every model in Section 4.1. ECE in Section 4.1 uses ten equal-width probability bins, weighted by their proportions of pooled held-out predictions across all fifteen runs. The adaptive check uses ten equal-frequency bins defined by prediction quantiles, collapsing duplicate boundaries and keeping tied probabilities together, with the same pooling and weighting.

Early stopping: Both the deep baselines and XGBoost stop on validation AUC computed on the fold's 10% validation students: patience 10 epochs for the deep models, 20 boosting rounds for XGBoost. The parameters restored for test scoring are those of the epoch or boosting round with the best validation AUC. The validation partition is used for early stopping, for the permutation-importance comparator, and for the rankings that feed the ablation experiments; it is never used for test-set reporting. The diagnostic figure in Section 4.1 supplies representative learning curves from additional diagnostic refits of XGBoost and DKT on rebuilt 2009 (seed 0, fold 0), using the same configuration; these refits do not replace the reported performance runs.

### 3.6 Statistical Reporting and the Unit of Analysis

The submitted manuscript reported 95% confidence intervals computed across the fifteen fold-by-seed runs as though those were fifteen independent observations. They are not. As Section 3.5 makes explicit, the five folds are overlapping subsamples of one student population, so the runs share both training and test students; treating them as independent understates uncertainty, which is the most likely explanation for intervals as narrow as ± 0.001 on ASSISTments 2012. This is a known hazard of resampled cross-validation [65, 66].

Uncertainty is therefore reported at the level of the sampling unit the design varies, which is the student. For each model, held-out predictions from every fold within a seed are pooled and a nonparametric bootstrap is taken over students, resampling students with replacement and recomputing the metric over all their interactions, for 1,000 replicates, with the reported interval running from the 2.5th to the 97.5th percentile, averaged over the three seeds. Resampling students rather than interactions respects the clustering of interactions within students and the temporal autocorrelation it induces. Bootstrapping within each seed rather than across all fifteen runs matters for the same reason: pooling first would enter each student roughly three times, since each seed draws five independent 20% test samples. Run-to-run spread across folds and seeds is reported separately as a standard deviation. For the ablation experiments, dispersion across the ten random draws is reported alongside the mean.

Reproducibility: Global seeding (random, numpy, torch, CUDA deterministic), pinned dependency versions, and provenance recording of the git commit, GPU model, CUDA version, and library versions for every run. The repository named in the Code Availability statement contains the preprocessing code, the feature-generation utilities, the model configuration files, the evaluation and figure scripts, the random seeds, dataset acquisition instructions, and the commands that reproduce each table and figure in this manuscript.

### 3.7 What These Evaluations Estimate

A student-wise random split estimates one thing: how well a model transfers to students who were not seen during training but who are drawn from the same cohort, the same platform, and the same period. It does not estimate how well a model transfers to a later cohort, to a subsequent academic term, or to interactions that occur after the training data ends, because training and test interactions are interleaved in time. Knowledge tracing is often motivated by exactly that prospective use, so the distinction is stated rather than left implicit. This manuscript's claims are scoped to unseen students from the same population. Where the submitted version implied prospective deployment performance, the text has been narrowed; a forward-in-time evaluation on ASSISTments 2012, whose wall-clock timestamps make one well defined, is identified in Section 6 as the appropriate next step rather than being claimed here.

## 4. Results

### 4.1 Predictive Accuracy (RQ1)

Predictive performance across models and datasets is reported in Table 5.

Table 5. Predictive performance by information set. AUC with 95% student-level bootstrap intervals over five folds and three seeds

| **Model** | **Information set** | **ASSISTments 2009 (rebuilt)** | **ASSISTments 2012** |
|---|---|---|---|
| *Identifiers and correctness only* | | | |
| DKT | `skill_id`, correct | 0.720 [0.712, 0.728] | 0.700 [0.697, 0.703] |
| SAKT | `skill_id`, correct | 0.693 [0.684, 0.701] | 0.680 [0.677, 0.684] |
| XGBoost (restricted) | derived from `skill_id`, correct | 0.717 [0.708, 0.725] | 0.697 [0.693, 0.700] |
| *Plus item identity* | | | |
| AKT | + `problem_id`, Rasch | 0.703 [0.695, 0.711] | 0.690 [0.687, 0.694] |
| SimpleKT | + `problem_id`, Rasch | 0.701 [0.693, 0.709] | 0.690 [0.687, 0.694] |
| *Plus temporal gaps* | | | |
| DKT-Forget | + elapsed gaps | 0.719 [0.711, 0.728] | 0.704 [0.701, 0.707] |
| *All behavioral features* | | | |
| SAKT + features | all behavioral | 0.704 [0.695, 0.712] | 0.688 [0.685, 0.691] |
| DKT + features | all behavioral | 0.723 [0.715, 0.731] | 0.708 [0.705, 0.711] |
| Logistic Regression | all behavioral | 0.754 [0.747, 0.761] | 0.756 [0.754, 0.758] |
| XGBoost (prediction-time) | no `response_time_log` | 0.775 [0.769, 0.782] | 0.771 [0.769, 0.773] |
| XGBoost | all behavioral | 0.786 [0.779, 0.792] | 0.777 [0.775, 0.779] |

Fig. 2 compares the predictive performances of the feature-based and deep KT models across both datasets.

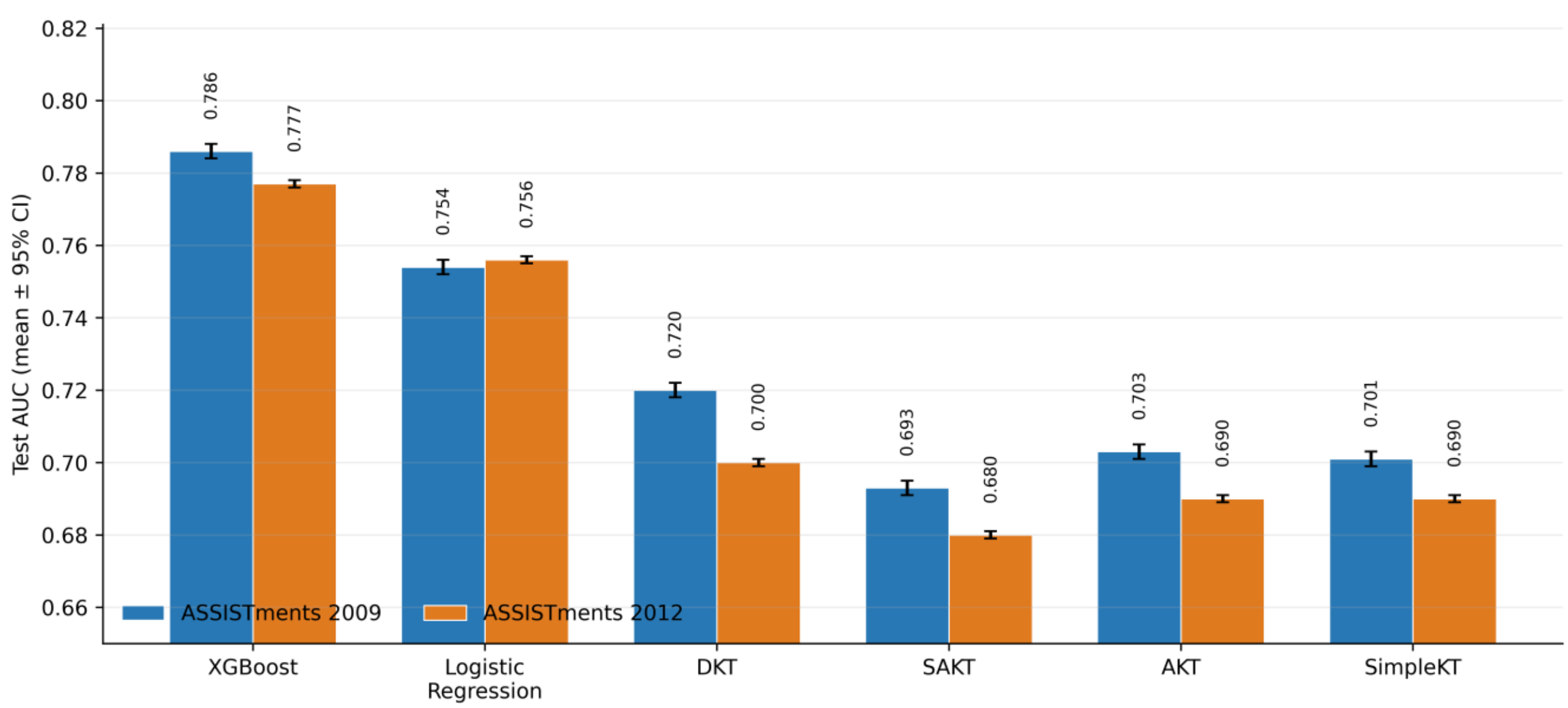

**Fig. 2** Predictive performance by dataset

RQ1 Answer: Table 5 groups every model by the information it receives. The reciprocal control is the more informative of the two directions: restricting XGBoost to quantities derivable from the identifier-and-correctness stream that DKT and SAKT consume brings it to 0.717 [0.708, 0.725] on ASSISTments 2009 against DKT's 0.720 [0.712, 0.728], and to 0.697 [0.693, 0.700] on ASSISTments 2012 against 0.700 [0.697, 0.703], similar on both, with overlapping marginal bootstrap intervals, and marginally below the recurrent baseline on ASSISTments 2012. These are marginal intervals rather than a paired test, so the appropriate reading is that the two are close, not that a difference has been excluded. The apparent advantage of the feature-based model is attributable to the information it receives, not to the model family. The enriched direction is more equivocal: supplying DKT with the same behavioral features raises it to 0.708 [0.705, 0.711] on ASSISTments 2012, an interval-separated gain, while on ASSISTments 2009 the intervals overlap almost entirely (0.723 [0.715, 0.731] against 0.720) and no gain can be claimed. DKT-Forget behaves similarly, and neither closes the distance to XGBoost, whose 0.786 and 0.777 are on par with published deep-KT results. The claim of a 7.7 to 9.7 point advantage is withdrawn.

Two caveats bound the second reading. The feature channel used here is deliberately minimal, the standardized features being projected by a learned linear map and added to the step embedding, which leaves width and capacity unchanged, and no model was tuned per architecture. A deep model given these signals through a richer mechanism, or tuned for them, might use them better. What the results support is that the deep baselines *as configured here* did not convert the additional information into the accuracy the gradient-boosted model obtains from it.

Uncertainty is reported as a student-level bootstrap interval, for the reasons set out in Section 3.6, with fold-and-seed variability reported separately. Sampling uncertainty about the student population was the larger of the two on both datasets, and it is the component the submitted run-level intervals understated by treating fifteen overlapping runs as independent observations.

Metrics beyond AUC are reported in Tables 6 and 7. They separate the models considerably more than AUC does: expected calibration error on ASSISTments 2012 ranges from 0.0006 for the restricted XGBoost to 0.0236 for AKT, an order of magnitude, while AUC spans only 0.68 to 0.78. With adaptive bins, ECE for restricted XGBoost, full XGBoost and AKT is 0.0008, 0.0023, 0.0236 on 2012 and 0.0033, 0.0028, 0.0105 on 2009, respectively. These are descriptive comparisons under the evaluated configurations, and their size depends on binning.

Table 6. Predictive metrics on ASSISTments 2009

| Model | 2009 ACC | 2009 RMSE | 2009 AUPRC | 2009 ECE |
|---|---|---|---|---|
| XGBoost | 0.746 | 0.413 | 0.865 | 0.0023 |
| XGBoost (prediction-time) | 0.740 | 0.419 | 0.857 | 0.0040 |
| XGBoost (restricted) | 0.712 | 0.438 | 0.807 | 0.0030 |
| Logistic Regression | 0.726 | 0.428 | 0.842 | 0.0067 |
| DKT | 0.713 | 0.439 | 0.808 | 0.0102 |
| DKT + features | 0.714 | 0.438 | 0.810 | 0.0079 |
| DKT-Forget | 0.713 | 0.439 | 0.806 | 0.0065 |
| SAKT | 0.697 | 0.449 | 0.785 | 0.0234 |
| SAKT + features | 0.705 | 0.444 | 0.794 | 0.0156 |
| AKT | 0.706 | 0.444 | 0.790 | 0.0107 |
| SimpleKT | 0.704 | 0.445 | 0.788 | 0.0131 |

Table 7. Predictive metrics on ASSISTments 2012

| Model | 2012 ACC | 2012 RMSE | 2012 AUPRC | 2012 ECE |
|---|---|---|---|---|
| XGBoost | 0.755 | 0.408 | 0.881 | 0.0023 |
| XGBoost (prediction-time) | 0.753 | 0.410 | 0.876 | 0.0023 |
| XGBoost (restricted) | 0.722 | 0.434 | 0.828 | 0.0006 |
| Logistic Regression | 0.742 | 0.417 | 0.867 | 0.0133 |
| DKT | 0.721 | 0.435 | 0.825 | 0.0089 |
| DKT + features | 0.724 | 0.432 | 0.831 | 0.0074 |
| DKT-Forget | 0.722 | 0.434 | 0.828 | 0.0077 |
| SAKT | 0.716 | 0.439 | 0.810 | 0.0099 |
| SAKT + features | 0.720 | 0.437 | 0.815 | 0.0062 |

| Model | 2012 ACC | 2012 RMSE | 2012 AUPRC | 2012 ECE |
|---|---|---|---|---|
| AKT | 0.720 | 0.437 | 0.816 | 0.0236 |
| SimpleKT | 0.720 | 0.437 | 0.816 | 0.0218 |

Figure 8 shows the additional diagnostic refits on rebuilt ASSISTments 2009 (seed 0, fold 0) described in Section 3.5.

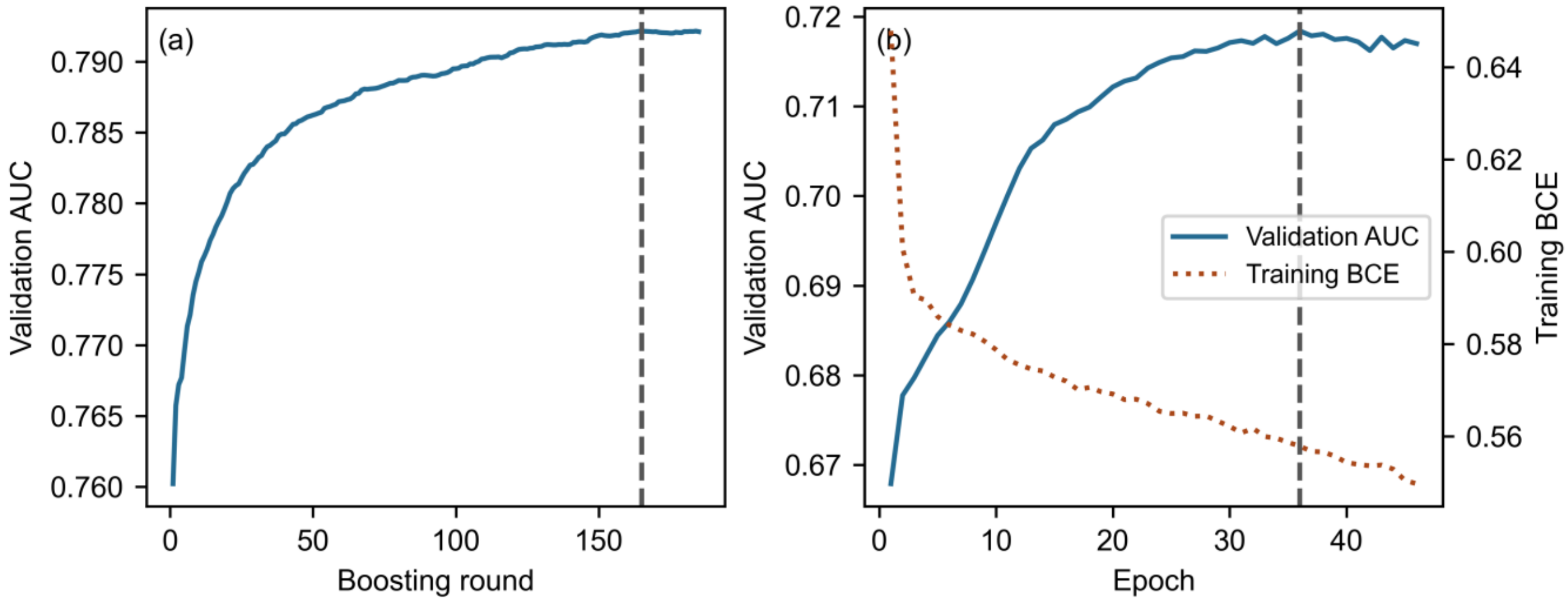

**Fig. 8** Diagnostic learning curves. (a) XGBoost validation AUC; (b) DKT validation AUC and training binary cross-entropy. Dashed lines mark best validation performance

### 4.2 Feature Importance and Stability (RQ2)

Global feature-importance values for both attribution methods are reported in Tables 8 and 9.

Table 8. Global SHAP importance on ASSISTments 2012. Mean |SHAP| ± sample SD over fifteen runs; path-dependent ranks

| Rank 2012 | Feature | Theme | 2012 Path dep. | 2012 Interv. |
|---|---|---|---|---|
| 1 | `problem_difficulty` | Difficulty | 0.594 ± 0.007 | 0.556 ± 0.008 |
| 2 | `prior_acc_skill` | Prior Mastery | 0.322 ± 0.004 | 0.280 ± 0.005 |
| 3 | `prior_acc_overall` | Prior Mastery | 0.214 ± 0.004 | 0.225 ± 0.006 |
| 4 | `rolling_acc_k` | Prior Mastery | 0.173 ± 0.002 | 0.173 ± 0.003 |
| 5 | `response_time_log` | Latency | 0.160 ± 0.004 | 0.175 ± 0.008 |
| 6 | `gap_since_skill` | Recency/ Spacing | 0.141 ± 0.002 | 0.143 ± 0.007 |
| 7 | `rolling_mean_response_time` | Latency | 0.103 ± 0.001 | 0.105 ± 0.003 |
| 8 | `skill_difficulty` | Difficulty | 0.076 ± 0.003 | 0.076 ± 0.002 |

| Rank 2012 | Feature | Theme | 2012 Path dep. | 2012 Interv. |
|---|---|---|---|---|
| 9 | `seq_position` | Engagement/ Position | 0.044 ± 0.001 | 0.058 ± 0.005 |
| 10 | `prior_attempts_skill` | Prior Mastery | 0.035 ± 0.001 | 0.040 ± 0.004 |
| 11 | `gap_since_last` | Recency/ Spacing | 0.032 ± 0.001 | 0.035 ± 0.002 |
| 12 | `attempts_in_session` | Engagement/ Position | 0.023 ± 0.001 | 0.024 ± 0.001 |
| 13 | `session_index` | Engagement/ Position | 0.020 ± 0.002 | 0.025 ± 0.003 |

Table 9. Global SHAP importance on ASSISTments 2009. Mean |SHAP| ± sample SD over fifteen runs; path-dependent ranks

| Rank 2009 | Feature | Theme | 2009 Path dep. | 2009 Interv. |
|---|---|---|---|---|
| 1 | `problem_difficulty` | Difficulty | 0.423 ± 0.007 | 0.406 ± 0.014 |
| 2 | `prior_acc_skill` | Prior Mastery | 0.360 ± 0.010 | 0.304 ± 0.010 |
| 3 | `response_time_log` | Latency | 0.291 ± 0.015 | 0.308 ± 0.024 |
| 4 | `prior_acc_overall` | Prior Mastery | 0.256 ± 0.009 | 0.263 ± 0.009 |
| 5 | `rolling_acc_k` | Prior Mastery | 0.213 ± 0.007 | 0.189 ± 0.012 |
| 6 | `skill_difficulty` | Difficulty | 0.140 ± 0.008 | 0.151 ± 0.010 |
| 7 | `gap_since_skill` | Recency/ Spacing | 0.126 ± 0.007 | 0.134 ± 0.008 |
| 8 | `rolling_mean_ response_time` | Latency | 0.119 ± 0.006 | 0.116 ± 0.008 |
| 9 | `gap_since_last` | Recency/ Spacing | 0.089 ± 0.008 | 0.099 ± 0.012 |
| 10 | `seq_position` | Engagement/ Position | 0.056 ± 0.011 | 0.068 ± 0.018 |
| 11 | `session_index` | Engagement/ Position | 0.038 ± 0.006 | 0.044 ± 0.009 |
| 12 | `prior_attempts_skill` | Prior Mastery | 0.035 ± 0.008 | 0.038 ± 0.008 |
| 13 | `attempts_in_session` | Engagement/ Position | 0.013 ± 0.004 | 0.016 ± 0.005 |

Fig. 3 presents the global TreeSHAP rankings and highlights the shared dominant signals and the secondary ordering differences between the datasets.

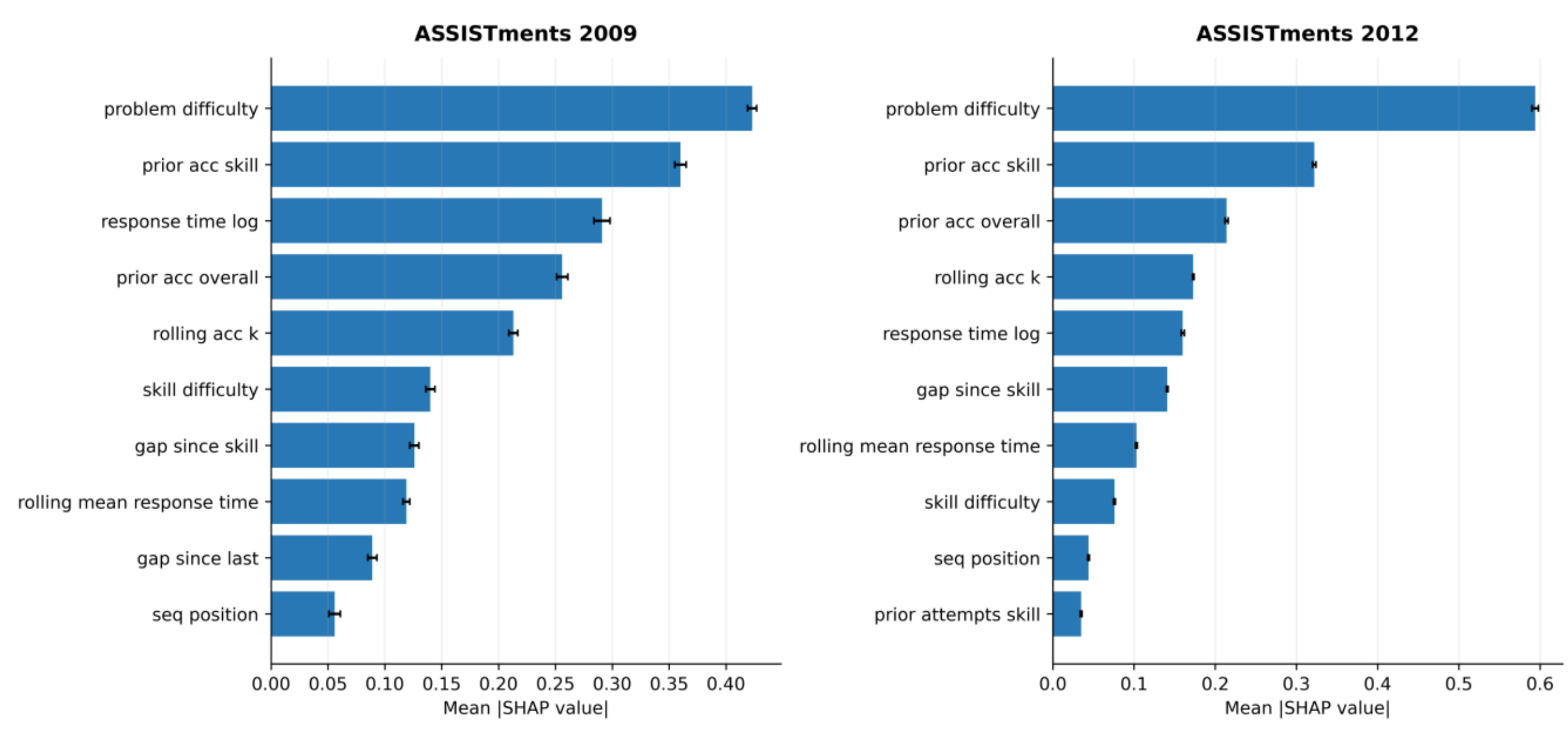

**Fig. 3** Global TreeSHAP feature importance

Cross-fold and inter-method stability results are reported in Table 10.

Table 10. TreeSHAP ranking stability across folds and conditioning schemes

| Dataset | Cross-Fold ρ (Path Dep.) | Cross-Fold ρ (Interventional) | Inter-Method ρ | Inter-Method τ |
|---|---|---|---|---|
| ASSISTments 2012 | 1.000 | 0.992 | 0.989 | 0.949 |
| ASSISTments 2009 | 0.994 | 0.996 | 0.996 | 0.983 |

Note. The 2009 inter-method cell (ρ = 0.996, τ = 0.983) averages correlations of within-seed fold-mean rankings. Correlating the grand-mean rankings in Tables 8 and 9 instead gives ρ = 0.9945 and τ = 0.9744; on 2012 it gives 0.989 and 0.949, the same to three decimals as the within-seed estimator. Both dataset cells use the same procedure.

RQ2 Answer: Rankings are highly consistent across folds and between conditioning schemes on both datasets. On ASSISTments 2012 the mean cross-fold Spearman correlation is 1.000 under path-dependent and 0.992 under interventional conditioning, and the two schemes agree at ρ = 0.989 (Kendall τ = 0.949). On the rebuilt ASSISTments 2009 corpus the corresponding values are 0.994, 0.996, ρ = 0.996 and τ = 0.983, higher than the contaminated corpus produced. Rankings computed with the gap features excluded remain stable (ρ = 0.995). The consistency of the ranking across partitions indicates that it is not an artifact of which students fall in a particular test split; it does not by itself establish that an explanation for any individual student is stable, a distinction developed in Section 5. Seed-to-seed ρ is 1.000 under both schemes on 2012, and 1.000 (path-dependent) and 0.996 (interventional) on 2009. As a check without averaging over folds, comparisons between seeds at matched fold indices give 1.000/0.990 on 2012 and 0.994/0.995 on 2009 (path-dependent/interventional).

Set overlap. On 2012, top-3 and top-5 Jaccard overlap is 1.000 across folds, seeds and schemes. On 2009, top-3 overlap is also 1.000 throughout; top-5 overlap is 1.000 across path-dependent folds, 0.956 across interventional folds, 1.000 across fold-averaged seed rankings under either scheme, and 0.978 between schemes on matched runs (minimum 0.667 for the two non-perfect comparisons). At matched fold indices, top-5 seed overlap is 1.000/0.956 on 2009 and 1.000/1.000 on 2012. Thus high full-ranking correlations coexist with occasional changes at the top-five boundary.

The cross-dataset path-dependent rank correlation is ρ = 0.929. `problem_difficulty` ranks first on both; the next four features on 2009 are `prior_acc_skill`, `response_time_log`, `prior_acc_overall` and `rolling_acc_k`, so a Latency feature ranks third. On 2012 the corresponding order is `prior_acc_skill`, `prior_acc_overall`, `rolling_acc_k` and `response_time_log`. The submitted manuscript reported a sharp divergence between the two datasets, with recency and spacing dominating ASSISTments 2009. That divergence does not survive the corpus rebuild and is withdrawn; Section 5 sets out what it actually was.

Prediction-time explanations. Re-explaining the saved twelve-feature models without `response_time_log` gives the following complete path-dependent rankings, in descending order. 2012: `problem_difficulty`, `prior_acc_skill`, `prior_acc_overall`, `rolling_acc_k`, `gap_since_skill`, `rolling_mean_response_time`, `skill_difficulty`, `seq_position`, `prior_attempts_skill`, `gap_since_last`, `attempts_in_session`, `session_index` (mean cross-fold ρ = 0.998; seed-to-seed ρ = 1.000). 2009: `problem_difficulty`, `prior_acc_skill`, `prior_acc_overall`, `rolling_acc_k`, `skill_difficulty`, `gap_since_skill`, `rolling_mean_response_time`, `gap_since_last`, `seq_position`, `prior_attempts_skill`, `session_index`, `attempts_in_session` (mean cross-fold ρ = 0.993; seed-to-seed ρ = 1.000). Difficulty and prior accuracy remain the leading signals. Top-3/top-5 fold overlap is 1.000/1.000 on 2012 and 1.000/0.844 on 2009.

### 4.3 Faithfulness (RQ3)

The top-k and random-k ablation results are reported in Table 11.

Table 11. Feature ablation: test AUC after retraining

| k | 2012 TreeSHAP | 2012 Random (mean ± SD, 10 draws) | 2012 Δ | 2009 TreeSHAP | 2009 Random (mean ± SD, 10 draws) | 2009 Δ |
|---|---|---|---|---|---|---|
| 1 | 0.733 | 0.767 ± 0.018 | 0.034 | 0.763 | 0.780 ± 0.009 | 0.017 |
| 2 | 0.724 | 0.766 ± 0.016 | 0.043[a] | 0.757 | 0.777 ± 0.009 | 0.021[a] |
| 3 | 0.715 | 0.757 ± 0.023 | 0.042 | 0.739 | 0.772 ± 0.013 | 0.032[a] |
| 5 | 0.647 | 0.750 ± 0.024 | 0.103 | 0.693 | 0.762 ± 0.018 | 0.069 |
| 7 | 0.632 | 0.708 ± 0.022 | 0.076 | 0.627 | 0.730 ± 0.012 | 0.103 |

| k | 2012 TreeSHAP | 2012 Random (mean ± SD, 10 draws) | 2012 Δ | 2009 TreeSHAP | 2009 Random (mean ± SD, 10 draws) | 2009 Δ |
|---|---|---|---|---|---|---|
| 9 | 0.568 | 0.688 ± 0.051 | 0.121[a] | 0.594 | 0.710 ± 0.035 | 0.116 |

[a] Δ is calculated from unrounded random and TreeSHAP means before rounding. The marked 2012 k = 2 and k = 9 and 2009 k = 2 and k = 3 cells therefore differ by 0.001 from subtraction of the displayed rounded columns.

Fig. 4 shows how the test AUC changes when the top-ranked SHAP features or randomly selected features are removed and the model is retrained.

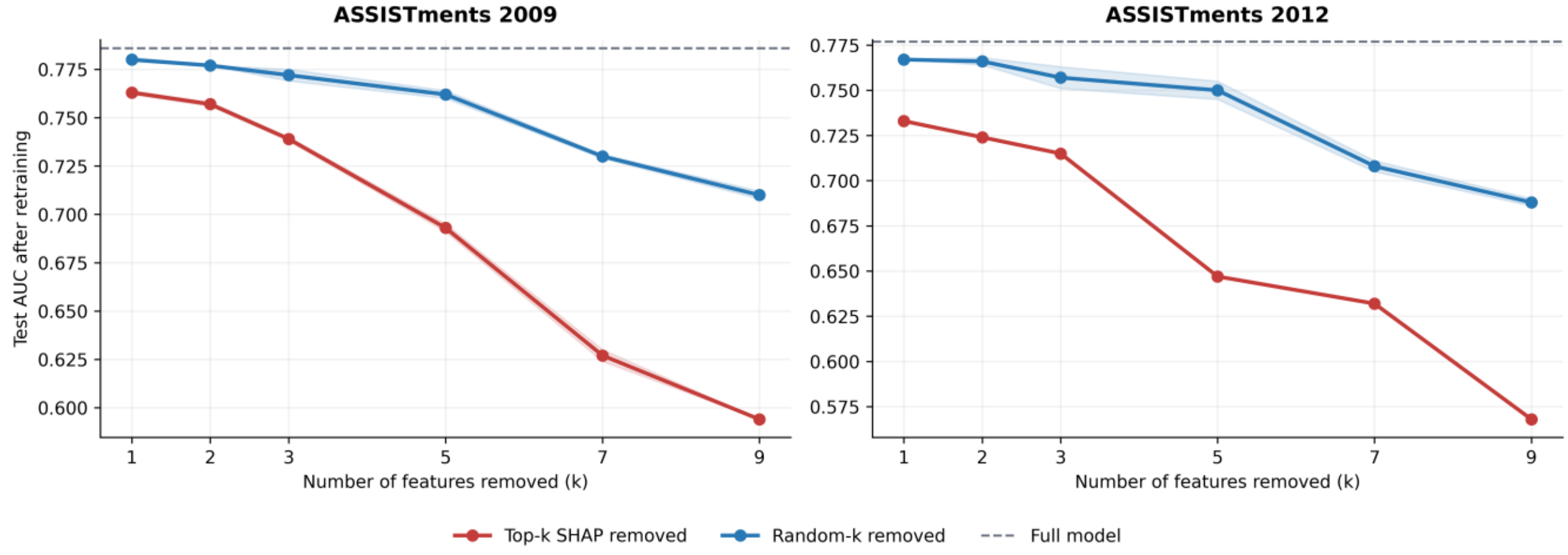

**Fig. 4** Faithfulness under feature ablation

Removing the features that TreeSHAP ranks highest degraded held-out AUC more than removing the same number of randomly chosen features, at every value of *k* tested, in all three seeds on both datasets, and both series decrease at every step. Table 11 reports the standard deviation across the ten random draws alongside each random mean, and that dispersion qualifies the small-*k* results in a way the submitted manuscript did not: at *k* = 1 the gap between ranked and random removal is 0.034 on ASSISTments 2012 against a draw-to-draw standard deviation of 0.018, and 0.017 on ASSISTments 2009 against 0.009, roughly two standard deviations in each case, whereas at *k* = 9 the gaps are 0.121 and 0.116 against 0.051 and 0.035. The faithfulness conclusion therefore rests on the whole series and on the larger removals, not on any single small-*k* cell.

### 4.4 Ablation Diagnostics

Monotonicity. The previously reported ASSISTments 2009 series rose from 0.805 at *k* = 1 to 0.816 at *k* = 3, and the *k* = 2 result the protocol specified was omitted in error; it is 0.817, which makes the reversal larger than reported. No reversal survives on the rebuilt corpus, where the series in Table 11 decreases at every step, so the reversal was an artifact of the leakage. Monotonic degradation is observed here, but it would not establish that redundancy among correlated features has been ruled out, and Section 5 does not offer it as such.

Reconciling the theme ablations with the top-*k* series. Tables 12 and 13 handle two properties. The themes are of unequal size, so the drop per feature removed is reported alongside the raw drop; and a theme is not a prefix of the ranking, so the two experiments remove different sets and need not order consistently. Under the corrected assignment the two datasets agree closely on the drop per feature, with Difficulty largest, then Prior Mastery, Latency, Recency/Spacing and Engagement/Position, an ordering that matches the feature-level ranking on both.

Table 12. Features and ranks removed in theme ablations

| Theme | Features removed | n | TreeSHAP ranks removed |
|---|---|---|---|
| Difficulty | `problem_difficulty`, `skill_difficulty` | 2 | 1, 8 |
| Prior Mastery | `prior_acc_skill`, `prior_acc_overall`, `rolling_acc_k`, `prior_attempts_skill` | 4 | 2, 3, 4, 10 |
| Latency | `response_time_log`, `rolling_mean_response_time` | 2 | 5, 7 |
| Recency/ Spacing | `gap_since_skill`, `gap_since_last` | 2 | 6, 11 |
| Engagement/ Position | `seq_position`, `attempts_in_session`, `session_index` | 3 | 9, 12, 13 |

Table 13. Theme ablation: raw AUC loss and loss per feature removed

| Theme | 2012 ΔAUC | 2012 ΔAUC per feature | 2009 ΔAUC | 2009 ΔAUC per feature |
|---|---|---|---|---|
| Difficulty | 0.056 | 0.028 | 0.041 | 0.021 |
| Prior Mastery | 0.044 | 0.011 | 0.036 | 0.009 |
| Latency | 0.007 | 0.004 | 0.013 | 0.006 |
| Recency/ Spacing | 0.004 | 0.002 | 0.004 | 0.002 |
| Engagement/ Position | 0.001 | 0.000 | 0.002 | 0.001 |

Comparison against non-random rankings. Table 14 reports the TreeSHAP ranking alongside XGBoost split gain and permutation importance under the identical remove-and-retrain procedure, with TreeSHAP and permutation importance computed on validation data and split gain obtained from the fitted model. TreeSHAP yields lower AUC than random removal at every tested k on both datasets. This does not hold uniformly for all three rankings: on 2009, split gain gives 0.784 against random removal at 0.780 ± 0.009 for k = 1, and 0.776 against 0.777 ± 0.009 for k = 2. The latter difference is small relative to the random-draw dispersion. The rankings are broadly competitive, and they often select the same features, so the greater part of the effect is that a reasonable importance ranking beats chance rather than anything specific to Shapley values. Where they diverge, TreeSHAP is usually but not always the more faithful. It is competitive with established alternatives under this test and is not uniquely faithful.

Table 14. Ranking comparison: test AUC after retraining. Lower is more faithful; random baseline in Table 11

| k | 2012 TreeSHAP | 2012 split gain | 2012 permutation | 2009 TreeSHAP | 2009 split gain | 2009 permutation |
|---|---|---|---|---|---|---|
| 1 | 0.733 | 0.733 | 0.733 | 0.763 | 0.784 | 0.763 |
| 2 | 0.724 | 0.724 | 0.724 | 0.757 | 0.776 | 0.748 |
| 3 | 0.715 | 0.712 | 0.711 | 0.739 | 0.750 | 0.739 |
| 5 | 0.647 | 0.647 | 0.668 | 0.693 | 0.693 | 0.705 |
| 7 | 0.632 | 0.632 | 0.586 | 0.627 | 0.633 | 0.630 |
| 9 | 0.568 | 0.596 | 0.568 | 0.594 | 0.594 | 0.598 |

Note. The fourth ranking counted in Section 3.5 is test-derived TreeSHAP. Its AUC series equals the validation-derived TreeSHAP series at every reported k for both datasets and every seed in the saved outputs, so those two rankings share the TreeSHAP columns. Split gain is extracted from the training-fitted model; the two other reported rankings are evaluated on validation data.

Directionality. Mean absolute SHAP values are unsigned and carry no information about the direction in which a feature moves a prediction, so they cannot support statements of the form *longer gaps predict failure*. Signed SHAP summaries and per-feature dependence plots are therefore reported in Figs. 5 and 6. They support the directional readings the Discussion relies on, with one qualification. Longer time since the student last practiced *the same skill* lowers the predicted probability of a correct response on both datasets, which is the direction the forgetting literature would predict. Time since the student's previous interaction of any kind does not behave consistently: it lowers the prediction on ASSISTments 2009 and raises it on ASSISTments 2012. Longer response latency lowers the prediction on both. Higher recent accuracy and higher item easiness raise it, as expected. Every directional statement in Section 5 is now either supported by one of these panels or marked as a hypothesis.

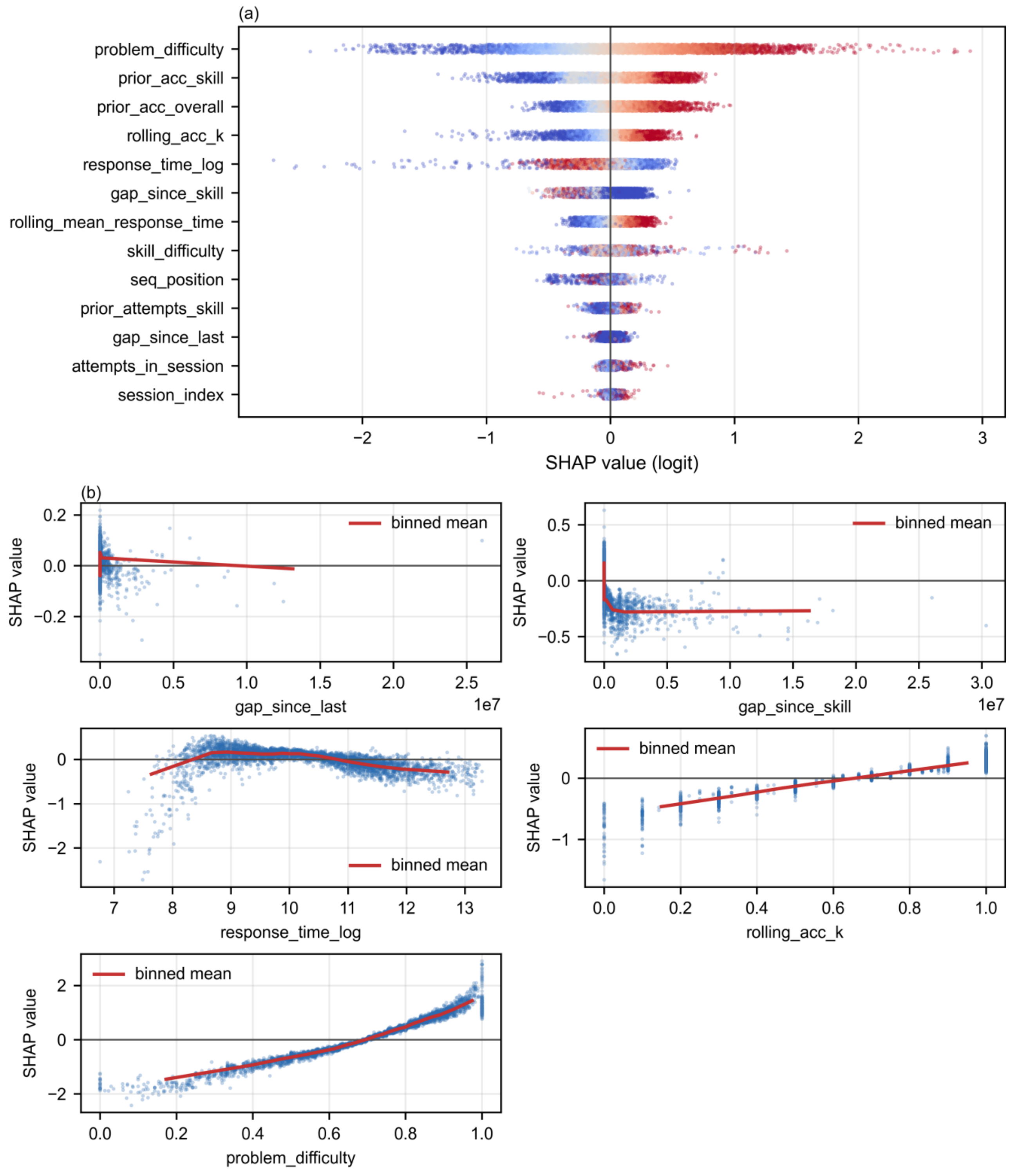

**Fig. 5** Signed SHAP values and dependence on ASSISTments 2012. (a) Signed attributions, blue to red for low to high values; (b) dependence plots with binned means

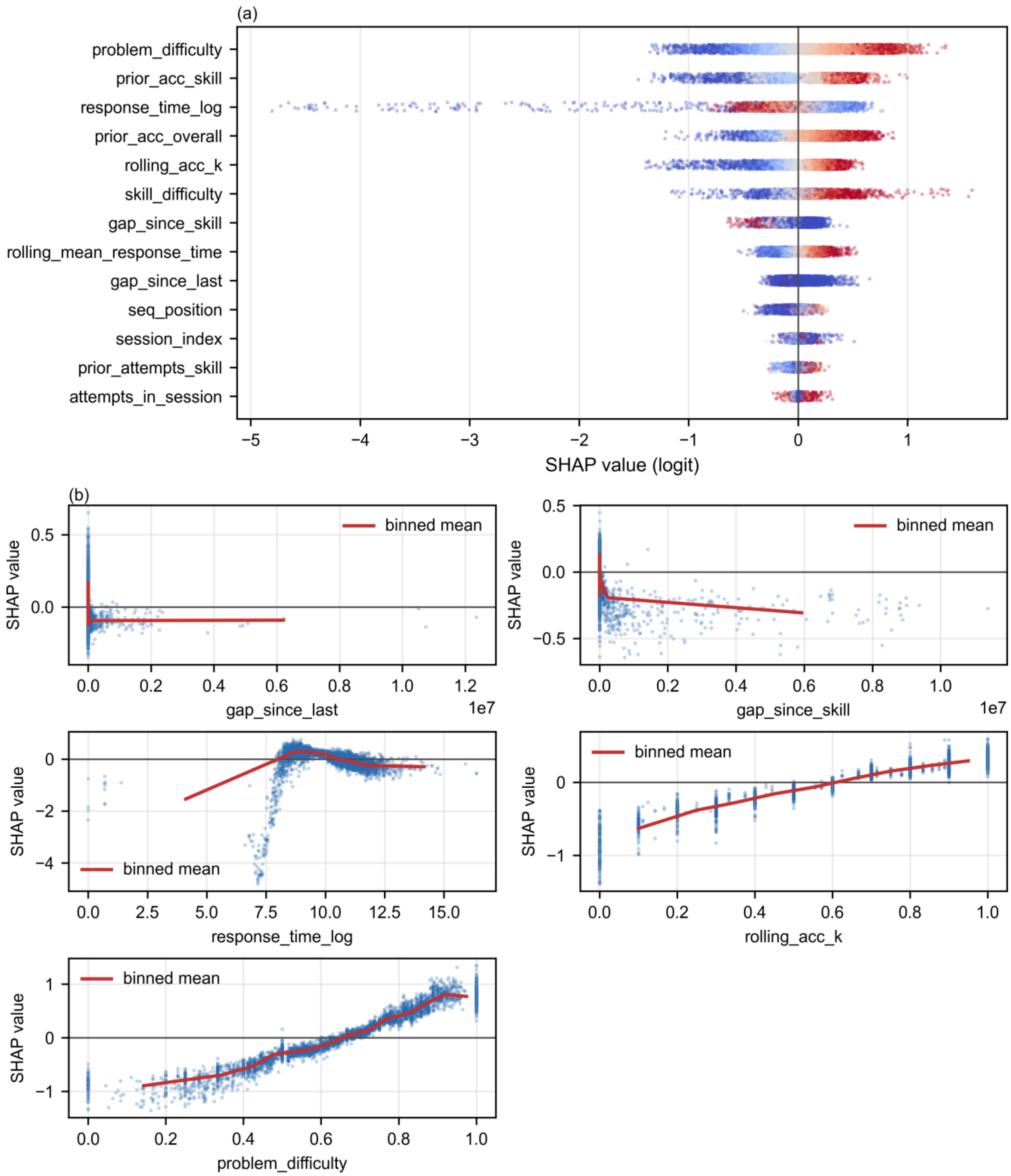

**Fig.** 6 Signed SHAP values and dependence on ASSISTments 2009. (a) Signed attributions, blue to red for low to high values; (b) dependence plots with binned means

RQ3 Answer: Removing the features TreeSHAP ranks highest degrades held-out performance more than removing an equal number at random, at every value of *k* tested, in all three seeds on both datasets, and theme-level ablation orders the themes consistently with the feature-level ranking. Split-gain and permutation rankings behave comparably under the identical procedure, so the result supports the weaker of the two available claims: a reasonable importance ranking is faithful in this sense, rather than the TreeSHAP ranking being uniquely so. These results establish that the highly ranked features influence model predictions under the stated ablation procedure; they do not establish that those features cause student performance.

### 4.5 Case Studies (Local Explanations)

Three students from ASSISTments 2012 illustrate how a local attribution decomposes an individual prediction. They were selected to span a range of predicted probabilities, and they carry a specific caveat: this study evaluated the stability and faithfulness of *global* feature rankings, not whether an individual student's local explanation is stable under resampling or faithful under ablation. A global ranking that does not move across partitions does not entail that the attribution for any particular student is reliable. The readings offered below are therefore hypotheses about what the model has learned to respond to, not validated statements about these students and not recommendations for instructional action; Section 6 identifies local stability and local faithfulness as necessary work before explanations of this kind are put in front of instructors.

#### *Struggling Student*

This student had 0% correct answers across 21 test interactions. The predicted P(correct) was 0.108. Local SHAP: `rolling_acc_k` = −1.254, `prior_acc_skill` = −1.221 (both pushing toward failure). Reading of the model: the two features carrying the largest negative attribution are both measures of the student's earlier accuracy, so the low predicted probability is attributable to the model's response to that history rather than to item difficulty or latency.

#### *High Performer*

This student had 100% correct answers across 32 test interactions. The predicted P(correct) = 0.963. Local SHAP: `problem_difficulty` = +1.245 (easy item), `prior_acc_overall` = +0.852, `rolling_acc_k` = +0.309. Reading of the model: the largest positive attribution is the item's encoded easiness, with the student's accumulated accuracy contributing further positive attribution. The prediction is consistent with a strong student encountering an item the training data records as widely answered correctly.

#### *Improving Student*

This student had 40% overall correctness, an improving trajectory, and 35 interactions. The predicted P(correct) was 0.544. SHAP shows competing signals: `problem_difficulty` = −0.815 (harder item) offset by `prior_acc_skill` = +0.356 (growing mastery). Reading of the model: the attributions carry opposite signs, with item difficulty pushing the prediction down and skill-level accumulated accuracy pushing it up, which is why the prediction sits near the decision boundary.

The local attributions for these three students are shown in Fig. 7, which reports what the model responded to rather than what an instructor should do.

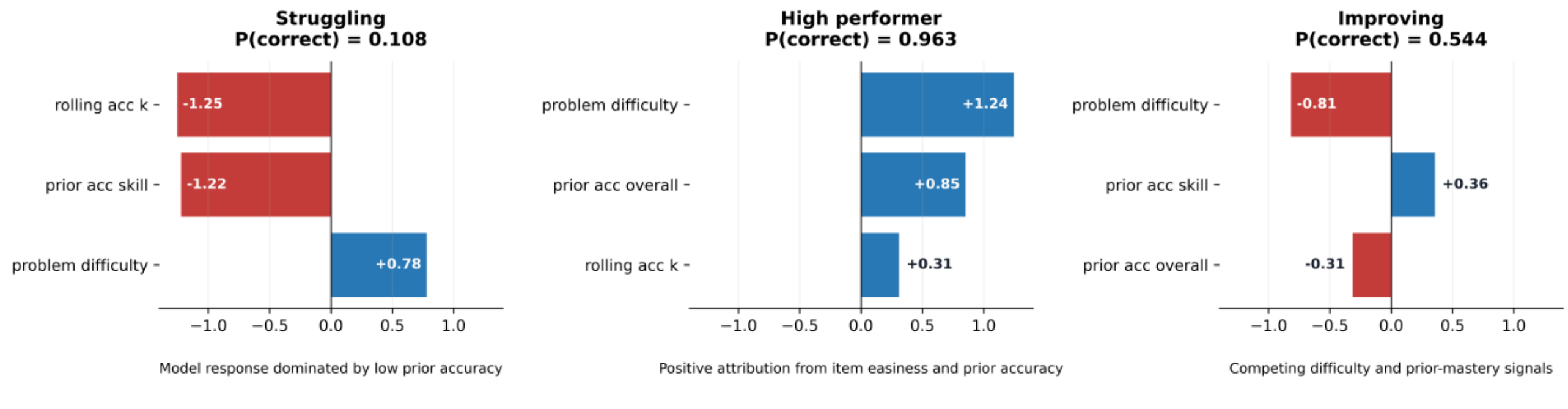

**Fig. 7** Local SHAP explanations for three students

## 5. Discussion

### Explanation Without an Accuracy Penalty

That a feature-based model matches deep KT on these benchmarks is consistent with prior findings [10, 11] and is not by itself novel. What this design adds is the reciprocal control, which locates why. Confined to what the deep baselines can derive from the identifier-and-correctness stream, the feature-based model performs similarly, its bootstrap interval overlapping theirs: 0.717 against 0.720 on ASSISTments 2009, and 0.697 against 0.700 on ASSISTments 2012. The difference reported in the submitted manuscript was a difference in information, not in model family, and no claim about the relative merits of the two families survives the control. The choice between them can therefore be made on grounds other than accuracy, among them whether the model admits exact, feature-level attribution. Under the shared, untuned configurations, restricted XGBoost has the lowest ten-bin equal-width ECE on ASSISTments 2012 (0.0006, against 0.0236 for AKT). The adaptive-bin comparison in Section 4.1 retains the ordering of these two models; it does not establish an architecture-level calibration advantage after tuning.

Three distinctions matter here, because the submitted manuscript blurred them. *Intrinsic interpretability* is a property of a model whose decision function can be inspected directly, as with a logistic regression's coefficients; a boosted ensemble of three hundred trees does not have it, and this manuscript no longer describes XGBoost as interpretable. *Post-hoc explainability* is supplied by a separate procedure applied to a fitted model, which is what TreeSHAP provides. Within it, *global feature attribution* summarizes which inputs a model relies on across a population and is what RQ2 and RQ3 examine, while *local explanation* decomposes a single prediction and is what Section 4.5 illustrates. The advantage claimed here is not intrinsic interpretability. It is an input space of named, pedagogically meaningful quantities over which exact Shapley values are computable and removal experiments are well defined, so the explanations can be *audited* in a way a deep model's latent state cannot.

### Stability as a Trustworthiness Criterion

High cross-fold agreement addresses the concern that a feature-importance ranking might depend on which students happen to fall in the test partition. Because each fold draws a different sample of students, agreement across folds indicates that the ranking reflects regularities in the population rather than the composition of one sample. Two limits should be read alongside it. The folds overlap, as Section 3.5 makes explicit, so cross-fold agreement is not agreement across independent samples; and stability of a global ranking is not stability of an individual explanation. Whether the attribution computed for one student would be reproduced under a different partition remains untested, and Section 6 identifies it as the most consequential piece of unfinished work.

The two conditioning schemes give similar rankings: inter-method Spearman agreement is $\rho = 0.989$ on ASSISTments 2012 and $\rho = 0.996$ on rebuilt ASSISTments 2009. This supports consistency of the ranking under these two choices. It does not remove correlations among behavioral features (Section 3.3). Interventional conditioning redistributes importance among correlated features, so agreement is weaker for attribution magnitudes. The stability claim therefore concerns feature order, not identical importance values.

### Data-Integrity and Attribution Corrections

Rebuilding ASSISTments 2009 materially changed its explanation results. Duplicate multi-skill rows leaked the current label into historical accuracy features, inflating `rolling_acc_k` in particular, because that feature reads the recent-correctness window into which a duplicated row deposits its own label. Its mean absolute attribution falls from 1.384 on the contaminated corpus, first of thirteen, to 0.213 on the rebuilt one, fifth of thirteen, and the Recency/Spacing theme drop on ASSISTments 2009 falls from 0.080 to 0.005. On the rebuilt corpus `problem_difficulty` ranks first on both datasets, so the cross-dataset divergence reported in the submitted manuscript disappears. That finding, the argument for per-deployment characterization of feature importance that rested on it, and any prescription for spaced review drawn from it are all withdrawn.

Two further corrections belong with it. Because `rolling_acc_k` measures recent correctness rather than the distribution of practice over time, it was reassigned from Recency/Spacing to Prior Mastery, which the correlation structure supports ($\rho = 0.64$ with `prior_acc_overall` and $\rho = 0.57$ with `prior_acc_skill` on ASSISTments 2012, against $\rho = -0.06$ and $\rho = -0.02$ with the two elapsed-gap features), and the theme analyses were repeated under the new mapping. Separately, an initial row-level leave-one-out difficulty encoding made each value a deterministic function of that row's own label, inflating `skill_difficulty` to roughly 36% of attributed importance and cutting held-out AUC on the affected folds from approximately 0.77 to approximately 0.63; it was replaced by the fold-internal cross-fitting of Section 2.4. Removal-based diagnostics exposed both defects, which shows that explanation validation can detect pipeline faults that aggregate accuracy does not reliably reveal, though it does not establish that the attribution method is uniquely reliable.

## Pedagogical Implications

The signed-attribution results suggest hypotheses for future instructional evaluation rather than validated recommendations. Learning analytics dashboards [39] report aggregate metrics such as pass rates and time on task, which describe what is happening but not what a model responds to; a SHAP decomposition could report which inputs moved a particular prediction, and instructors are known to prioritize explanations that map to actions they can take [40]. Longer skill-specific gaps were associated with lower predicted correctness and could motivate testing spaced review; prior-mastery signals could motivate testing remediation or scaffolding; and difficulty signals could inform item sequencing against the zone of proximal development [41]. General elapsed time was inconsistent across the two datasets and should not be read pedagogically. Latency attributions cannot be equated with gaming, because no validated detector [42] and no ground-truth labels were available here. Because the decomposition is per interaction step, it is at the grain size at which tutoring feedback is effective [43]. Whether instructors read such explanations correctly, and whether acting on them improves learning, remains untested; transparency alone does not improve human decisions without attention to interface design and training [44].

## Threats to Validity

Construct validity. This study assumes that SHAP values constitute meaningful explanations of model behavior. While Shapley values satisfy desirable axiomatic properties (efficiency, symmetry, linearity, null player) [45], they explain the model's computation, not the underlying causal process generating student data. A feature may be important to the model because it correlates with a true causal driver rather than being causal itself. The explanations should be read as statements that the model relies on these signals, not as statements that these factors cause student performance. This manuscript accordingly avoids the word *causal* as a descriptor of features, of attributions, or of the relationships they express; the observational design, with no randomization and no identification strategy, does not support causal inference.

Internal validity. The faithfulness test can be confounded if removing top-ranked features forces the refitted model onto previously redundant substitutes that partially recover the lost performance. The submitted manuscript claimed that observing monotonic degradation across increasing *k* mitigated this. That reasoning is withdrawn even though the observation now holds on the rebuilt corpora the ablation series decreases at every step on both datasets, but monotonicity is not evidence that redundancy has been ruled out, and the submitted manuscript was wrong to offer it as such. What the design does support is the comparative statement: under an identical remove-and-retrain procedure, the TreeSHAP ranking degrades held-out performance more than random removal of equal size, and Section 4.4 additionally reports how it compares with split-gain and permutation rankings. Redundancy among correlated behavioral features is a property of these data and limits what any removal-based criterion can establish about an individual feature in isolation.

Statistical conclusion validity. The submitted manuscript treated the fifteen fold-by-seed runs as fifteen independent observations. They are not independent: the folds are overlapping subsamples of one student population, so the runs share training and test students, and the resulting intervals were too narrow. Uncertainty is now reported by bootstrapping over students, as Section 3.6 describes, with run-to-run spread reported separately. A residual limitation remains interactions within a student are temporally autocorrelated, and while student-level resampling respects the clustering, it does not model that autocorrelation explicitly. A further limitation is that both datasets come from one platform, so the bootstrap describes uncertainty about the ASSISTments student population and not about students in general.

## 6. Limitations and Future Work

Both datasets originated from the ASSISTments platform [33] and share platform-level design choices despite differences in scale, temporal resolution and skill granularity, so the bootstrap intervals of Section 3.6 describe uncertainty about the ASSISTments student population rather than about students in general. The submitted version read a cross-dataset divergence in feature rankings as evidence of context dependence; that divergence did not survive the corpus rebuild, so per-deployment validation remains advisable on the general ground that these are two corpora from one platform. Replication on EdNet [46], Junyi Academy [47] or MOOCCubeX would strengthen the generalization claims.

The evaluation estimates transfer to unseen students from the same cohort and period and not transfer to a later cohort or to interactions occurring after the training window. A forward-in-time evaluation, in which training precedes testing chronologically, would estimate the latter and is well defined on ASSISTments 2012 because it carries wall-clock timestamps. It was not run here, and the manuscript's claims have been narrowed accordingly rather than extended by assumption.

Explanation-level validation was conducted at the global level. Whether an individual student's attribution is stable under resampling, and whether it is faithful under a local removal test, was not measured, and because local explanations are what an instructor would see, establishing those properties is the most consequential piece of unfinished work identified here.

Implementing the deep baselines directly rather than adopting pyKT [60] provided the access to intermediate activations that the attribution work required, but it means the baselines are not the community-benchmarked implementations, and their absolute numbers should be read as this study's implementations under a shared standardized configuration rather than as reference values for those architectures.

ASSISTments 2009 carries no wall-clock timestamps, so its gap features are positional proxies computed over an ordering key. A positional gap is not an elapsed interval, and no conclusion about spacing or forgetting in the sense used in the learning-science literature [30, 31] can be drawn from this dataset. Statements about spacing in this manuscript are therefore restricted to ASSISTments 2012.

A formal comparison between SHAP and deep-model attention/IG was excluded because the 5-theme mapping was too coarse for a meaningful rank correlation at $n = 5$. A finer-grained methodology (e.g., per-interaction attribution alignment) is required.

Features were designed a priori based on domain knowledge. Automated feature discovery or learned representations may capture additional signals, but at the cost of interpretability, which is the central design priority of this study.

This study found the global feature-importance rankings to be stable and faithful under the stated evaluation procedures; it did not test whether providing them to instructors improves learning outcomes. Future work should therefore extend to more diverse KT corpora, tune each architecture for a sharper accuracy comparison, develop finer-grained cross-model attribution, and test through controlled trials whether the identified signals translate into effective pedagogical interventions.

## 7. Conclusion

On the datasets tested, knowledge-tracing predictions can be accompanied by global feature-importance rankings whose stability and faithfulness have been measured rather than assumed, while retaining competitive predictive accuracy under the evaluated configurations. RQ1: an XGBoost model on the stated behavioral feature sets matched the deep KT architectures evaluated here and is on par with published results for these benchmarks; restricted to the information those baselines receive, it performs as they do. The apparent difference between the model families was a difference in the information supplied, and no claim of superiority is made.

RQ2: the global feature-importance ranking was highly consistent across folds, seeds and the two TreeSHAP conditioning schemes on both datasets. Local explanation stability was not evaluated.

RQ3: removing top-ranked features and retraining degraded held-out performance more than removing an equal number at random, at every *k* and in all seeds on both datasets, but split gain and permutation importance behave similarly under the same procedure, so TreeSHAP is competitive with the alternatives rather than uniquely faithful. The highly ranked features influence model predictions under the stated ablation procedure; they are not shown to cause student performance.

The practical implication is that a knowledge-tracing prediction can be decomposed into the contributions of named behavioral quantities, and that the decomposition can be audited for stability and faithfulness before it is relied upon. Whether such decompositions help instructors, and whether acting on them improves student outcomes, remains untested and is the necessary next step.

## Acknowledgements

Not applicable.

## Declarations

### Funding

This study did not receive any specific grants or funding.

### Competing Interests

The authors declare no competing financial or nonfinancial interests. This research was conducted independently and does not represent the views, opinions, or endorsements of the authors' respective employers.

### Ethical Approval

Not applicable.

### Consent to Participate

Not applicable.

### Consent to Publish

Not applicable.

### Data Availability

The publicly available ASSISTments 2009–2010 skill-builder dataset is available on Figshare at https://doi.org/10.6084/m9.figshare.25309000 [35]. The experimental pipeline downloads skill_builder_data.csv from the file identifier 44732737 in that record. The ASSISTments 2012-2013 dataset with affect predictions is available on Figshare at https://doi.org/10.6084/m9.figshare.25310431 [34]. The datasets were used subject to their applicable access and usage conditions. The processed summary results supporting the findings are included in the article and its accompanying materials.

### Code Availability

The complete source code is publicly available and was made available to the reviewers during this revision round. The repository contains the data preprocessing code, the feature-generation utilities, the model configuration files, the training and evaluation scripts, the explanation and ablation pipeline, the re-scoring and figure scripts, the random seeds, instructions for obtaining both datasets, and the commands that regenerate every table and figure in this manuscript. The archived release is deposited at https://doi.org/10.5281/zenodo.22715077.

### Author Contributions

P.P. contributed to the investigation and validation of the study and reviewed and revised the manuscript. A.M. conceived the study, developed the methodology and software, conducted the investigation and formal analysis, curated the data, prepared the visualizations, and drafted the manuscript. U.I. contributed to the study conception, supervised and coordinated the project, and reviewed and revised the manuscript. P.K.D.K. contributed to the methodology, software development, and validation, reviewed and revised the manuscript, and served as the corresponding author. All authors have read and approved the final manuscript and agree to be held accountable for their contributions.